\documentclass[10pt,twocolumn,letterpaper]{article}

\usepackage[english]{babel}
\usepackage{times}
\usepackage{epsfig}
\usepackage{subfig}
\usepackage{graphicx}
\usepackage{amsmath}
\usepackage{amssymb}
\usepackage{algorithm}
\usepackage{algpseudocode}
\usepackage{colortbl}
\usepackage{multirow}
\usepackage{xcolor}
\usepackage{booktabs}
\usepackage{titling}
\usepackage[T1]{fontenc}
\usepackage[pagebackref=true,breaklinks=true,letterpaper=true,colorlinks,bookmarks=false]{hyperref}
\title{Sample4Geo: Hard Negative Sampling For Cross-View Geo-Localisation}
\author{Fabian Deuser \hspace{4em} Konrad Habel  \hspace{4em} Norbert Oswald \\
University of the  Bundeswehr Munich \\
Institute for Distributed Intelligent Systems \\
Munich, Germany \\
{\tt\small fabian.deuser@unibw.de, konrad.habel@unibw.de, norbert.oswald@unibw.de}
}
\begin{document}
\date{}
\maketitle
\begin{abstract}
Cross-View Geo-Localisation is still a challenging task where additional modules, specific pre-processing or zooming strategies are necessary to determine accurate positions of images. Since different views have different geometries, pre-processing like polar transformation helps to merge them. However, this results in distorted images which then have to be rectified. Adding hard negatives to the training batch could improve the overall performance but with the default loss functions in geo-localisation it is difficult to include them.
In this work, we present a simplified but effective architecture based on contrastive learning with symmetric InfoNCE loss that outperforms current state-of-the-art results. Our framework consists of a narrow training pipeline that eliminates the need of using aggregation modules, avoids further pre-processing steps and even increases the capacity of generalisation of the model to unknown regions. We introduce two types of sampling strategies for hard negatives. The first explicitly exploits geographically neighboring locations to provide a good starting point. The second leverages the visual similarity between the image embeddings in order to mine hard negative samples. Our work shows excellent performance on common cross-view datasets like CVUSA, CVACT, University-1652 and VIGOR. A comparison between cross-area and same-area settings demonstrate the good generalisation capability of our model.

\end{abstract}
\section{Introduction}
\begin{figure}[t!]
  \centering
    \includegraphics[width=0.7\columnwidth]{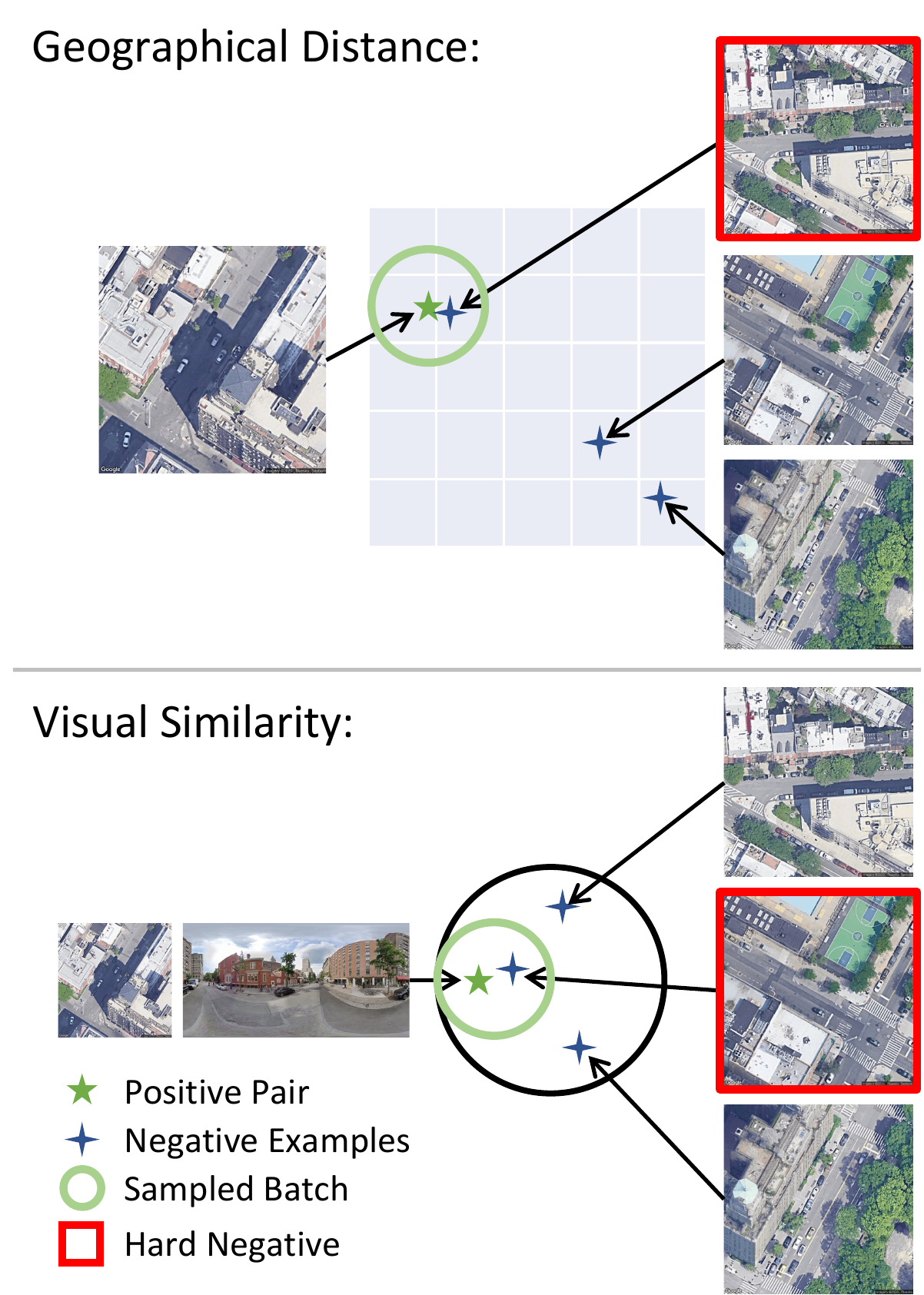}
  \caption{Our two sampling strategies. The first is based on the geographical distance between the satellite images. The second uses the cosine similarity between the street-view and satellite view embeddings to find hard negatives within a margin.}
  \label{fig:samplingStrats}
\end{figure}
Determining a geo-location from images without metadata is one of the puzzles yet to be solved in the computer vision community. Solving this problem can help in areas such as agriculture and automotive. For example, a robotic agent in agriculture for spraying fertiliser requires a high precision location. This can be achieved with a real-time kinematic (RTK) GPS, but these sensors are expensive and short-time signal outages can obstruct workflows. Therefore aerial image based localisation in these highly repetitive environments can further enhance positioning~\cite{Chebrolu2019robotlocal}. Another challenge in cities is the so-called urban canyon effect, which blocks signals such as GPS or reduces their accuracy. 
Especially in large cities, GPS signals are noisy due to tall buildings. The evaluation of 250,000 driving hours in New York City traffic by Brosh et al. showed an error of 10 meters in 40\% of GPS signals. Therefore a computer vision solution based on image retrieval was proposed~\cite{brosh2019accurate} to correct these signals.

While classic approaches sought to do this with visual clues such as sun position and the resulting shadows~\cite{cozman1995sextant,lalonde2010sky, Imran2010gps} or weather~\cite{Jacobs2008, Jacobs2011}, current approaches are increasingly focusing on image retrieval based on deep learning~\cite{arandjelovic2016netvlad, Lin2015, liu2019lending}. In cross-view image retrieval~\cite{Workman_2015_CVPR_Workshops,workmann2015cvusa,zheng2020university, zhu2021vigor}, different views of an image, e.g. a ground image and a satellite image, have to match to determine the searched position. The ground image queries are compared with a database of satellite images with known geographical positions.

Previous approaches mostly used CNN based methods~\cite{workmann2015cvusa, hu2018cvm, cai2019SiamFCANet, shi2020optimal,Wang2022LearningCG,wang2021each, shi2019safa, shi2020looking}, while present research is mainly focused on the Transformer or MLP Mixer architecture as a backbone for geo-localisation~\cite{yang2021l2ltr, zhao2022mutual, zhu2022transgeo,zhu2023simple}. In the latter case, reasonable performance can only be achieved if the weights between the two view-specific encoders are not shared, resulting in larger models. Furthermore, the polar transformation is often utilised to bridge the geometrical gap between the views~\cite{shi2019safa}. Triplet loss as a standard loss in cross-view geo-localisation uses only one negative example per batch and is prone to model collapsing when hard negatives are used~\cite{wu2017sampling}. 

In this paper, we present a weight-sharing Siamese CNN that learns class-agnostic embeddings based on the InfoNCE loss~\cite{oord2018representation,he2020momentum,chen2020simple} and demonstrate the advantage of the CNN architecture for our approach. We use a technique from multimodal pre-training~\cite{radford2021learning} to calculate the loss symmetrically to further aid the understanding of the different domains in the viewpoints. Hard negative examples, i.e. examples that are hard to distinguish for the model from positives examples, in deep metric learning are a key ingredient to achieve superior performance, thus we present two sampling methods. In the early epochs, we leverage GPS coordinates to contrast close geographic neighbours as a good initialisation for our second sampling. In later epochs based on a similarity metric, such as cosine similarity, we collect visually similar samples for the batch construction, to focus on hard negatives examples. 
To summarise our contribution in this work:
\begin{itemize}
    \item We show the superior performance of our contrastive training with the symmetric InfoNCE loss whilst using a single encoder for both views.  
    \item We propose GPS-Sampling as a task-specific sampling technique for hard negatives at the beginning of the training.
    \item We present Dynamic Similarity Sampling (DSS) to select hard negatives during training based on the cosine similarity between street and satellite views.
    \item Our framework consists of a simple training pipeline that eliminates the need of special aggregation modules or complex pre-processing steps, whilst outperforming current approaches in performance and generalisation ability. 
\end{itemize}
\section{Related Work}
One of the first works by Workman et al.~\cite{workmann2015cvusa} showed that CNN extracted features are far superior to hand-crafted features. In their work, they also presented CVUSA which is nowadays one of the main benchmarks for geo-localisation. For training their two-stream CNN as a Siamese Network~\cite{chopra2005simiaritymetric, taigman2014deepface} a simple L2 target was used to reduce the distance between view specific features. In following work Zhai et al.~\cite{zhai2017predicting} used noisy semantic labels for the aerial view and minimised the distance between these labels and a learned transformation for the street-view. In particular, this should better reflect the common semantic layout within the views. Vo et al.~\cite{vo2016localizing} introduced the soft-margin triplet loss as a standard for cross-view geo-localisation. The Triplet loss decreases the distance of a positive example to an anchor and increases the distance to a negative example. Subsequent work from Hu et al.~\cite{hu2018cvm} included the NetVLAD-layer~\cite{arandjelovic2016netvlad} in their architecture to further enhance the global description generation. The outputed feature maps are aggregated based on a differential clustering instead of a simple average or max pooling. Because of the slow convergence of the soft-margin triplet loss they proposed a weighted soft-margin ranking loss where the distance between a positive and a negative sample is additionally scaled.

Initially attention was also paid to the orientation within the images, as in datasets such as CVUSA~\cite{workmann2015cvusa} the alignment is known a priori and can serve as an additional feature, as Liu et al.~\cite{liu2019lending} showed. To help the network better exploit directions, the number of input channels was expanded and colour-coded orientations, from top to bottom and left to right, were used for both street-view and satellite-view. Furthermore, they introduced CVACT as a dataset. 

In several subsequent publications~\cite{vo2016localizing, cai2019SiamFCANet, shi2020looking, hu2022beyondgeolocal}, the orientation was further used to create an auxiliary objective. The street-view image is shifted as an augmentation and the degree of shift has to be predicted. 

Since the geometry between the satellite image and the street-view differs, Shi et al.~\cite{shi2019safa} applied polar transformation to the satellite image. The resulting stretched image is similar to the domain of the street-view image, this is still a common pre-processing technique today. In addition, they extended their network with a Spatial-Aware Feature Aggregation (SAFA) module to pool important features from the feature maps in a learnable fashion. 



The polar transformation has the disadvantage that the distortion injects disturbances into the image. Toker et al.~\cite{toker2021coming} developed an approach that learns to remove these disturbances with the help of a GANs~\cite{goodfellow2020generative}. The street-view image was used as ground truth for the discriminator of the GAN. At inference time, the generator and discriminator were no longer used and only the latent representation is used to find similarities to the street-view. 

Yang et al.~\cite{yang2021l2ltr} introduced the use of a ResNet Backbone in combination with a Transformer for that task. The feature maps output by the CNN are provided with a positional encoding and then entered into the Transformer. 

Because the previous mentioned benchmarks are slowly saturating, a more challenging dataset named VIGOR was created by Zhu et al.~\cite{zhu2021vigor}.
While CVUSA, CVACT and VIGOR only use street-view to satellite images Zheng et al. created University-1652~\cite{zheng2020university} with UAV images instead of street-views. 


Previous approaches commit to a single step of prediction which can not be corrected, therefore TransGeo~\cite{zhu2022transgeo} and SIRNet~\cite{lu2022iterative} have emphasised additional refinement. In SIRNet, additional refinement modules are added to the CNN backbone. A Softmax-based decision process uses a variable number of modules, up to four. With TransGeo, an additional zoom step is performed using the attention map in its Transformer architecture. This results in smaller objects being viewed at a higher resolution. To increase the generalisation of their approach, TransGeo also uses Adaptive Sharpness-Aware Minimisation (ASAM)~\cite{kwon2021asam} to smooth the loss surface. ASAM significantly slows down the training process, because for all training data an additional forward pass through the network is needed.  
Since the geometric perspective is strongly shifted by the different viewing angles, GeoDTR~\cite{zhang2022cross} tries to extract additional geometric properties with a Transformer-based extractor. First, features are generated independently using a CNN view. However, since ground truth labels for these geometric features are missing, an additional counterfactual loss is used to contrast the output features of the CNN with those of the Geometric Extractor to ensure dissimilarity. A triplet loss is then calculated between the features of the view-dependent geometric extractors. 
An alternative architecture, namely SAIG-D, by Zhu et al.~\cite{zhu2023simple} utilises a MLP-Mixer~\cite{tolstikhin2021mlp}. They replaced the patch stem with a convolutional stem to support local feature learning. In addition they propose a feature aggregation module of its own, based on two fully connected layers with a GELU activation function and a following down projection. Similar to TransGeo they utilised Sharpness-aware minimisation (SAM)~\cite{foret2020sharpness} to further smooth the loss surface. 
\section{Methodology}
Deep metric learning for cross-view geo-localisation uses Triplet loss by default~\cite{schroff2015facenet}, with various extensions such as soft-margin triplet loss~\cite{vo2016localizing} or weighted soft-margin triplet loss~\cite{hu2018cvm}. Triplet loss aims to decrease the distance between a positive example and an anchor, while increasing the distance between a negative example and the anchor. To use Triplet loss, suitable triplets must be sampled beforehand. Based on this formulation, only one positive and the anchor is always contrasted with one negative example. Arandjelovic et al.~\cite{arandjelovic2016netvlad} already showed the advantage of using two negatives as quadruple loss instead of a triplet. Without the aforementioned extensions hard negatives in the Triplet loss result in a collapsed model~\cite{wu2017sampling}. With this in mind we designed our framework to leverage multiple hard negatives.
\begin{figure*}[t!]
  \centering
    \includegraphics[width=0.8\textwidth]{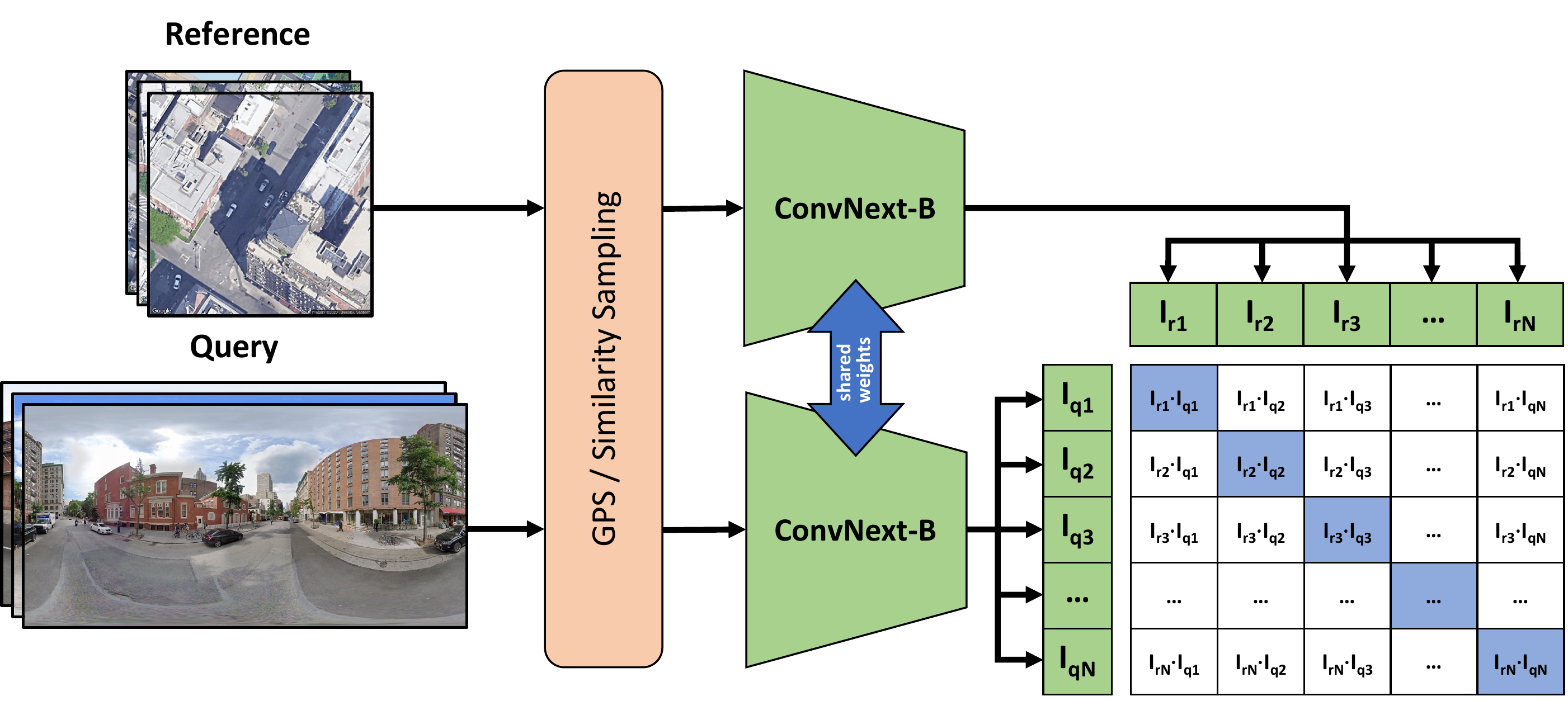}
  \caption{\textbf{Architecture overview of our approach.} We use an off-the-shelf ConvNeXt-B and mine hard negatives based on our GPS and visual similarity sampling. The InfoNCE loss is used in a symmetric fashion to learn discriminative features in both view directions.}
  \label{fig:ConvGeoArch}
\end{figure*}

\subsection{Symmetric InfoNCE Loss}
An alternative for contrastive learning that exploits all available negatives in the batch, as opposed to triplet loss, is the so-called InfoNCE~\cite{oord2018representation, radford2021learning} or NTXent loss~\cite{he2020momentum,chen2020simple}, see equation \ref{eq:infonce}.
\begin{equation}
\label{eq:infonce}
    \mathcal{L}(q,R)_{\text{InfoNCE}} = -\log \frac{\exp(q \cdot r_+/\tau))}{\sum_{i=0}^{R} \exp(q \cdot r_i/\tau))} 
\end{equation}
$q$ denotes an encoded street-view, the so-called query, and $R$ is a set of encoded satellite images called references. Only one positive $r_i$, namely $r_+$ matches to $q$. The InfoNCE loss uses the dot-product to calculate the similarity between query and reference images and is low when the query and the positive match are similar, and high when the negative $r_i$ are dissimilar to $q$. As loss function for the similarity between the views the cross-entropy is calculated. The temperature parameter $\tau$ is a hyperparameter that can either be learned~\cite{radford2021learning} or set to a static value.  
So far, InfoNCE loss has mostly been used in a non-symmetric fashion for unsupervised representation learning for images~\cite{he2020momentum,chen2020simple}. A symmetric formulation showed to be useful in multi-modal pre-training~\cite{radford2021learning} to bridge the gap between the modalities. Therefore we utilise this loss function in the same symmetric fashion to leverage the flow of information in both directions: satellite-view to street-view and vice versa. In the InfoNCE loss, a positive example is always contrasted with N-1 negative examples, where N denotes the batch size, thus delimiting many examples at once. But in cases where there are several positive examples, such as University-1652, this requires a custom sampler to prevent multiple positives for the same ground truth label in one batch. We provide an ablation study of the importance of symmetry in the InfoNCE loss and a comparison to the triplet loss in our supplementary material. 
\subsection{Model Architecture}
One of our main goals in this paper is to use an off-the-shelf architecture that does not require any special adaptations to the imagery in order to solve the geo-localisation problem in a meaningful way. Since recent publications have increasingly used Vision-Transformer~\cite{zhu2022transgeo,yang2021l2ltr,zhao2022mutual} or MLP mixer architectures~\cite{zhu2023simple}, we compare two architectures. For this purpose we used ConvNeXt~\cite{liu2022convnet} as state-of-the-art CNN and a Vision Transformer~\cite{dosovitskiy2020image}. We decide in favour of ConvNeXt on the basis of the results in section~\ref{sec:archchoice}. In line with previous approaches we use a Siamese Network as depicted in figure~\ref{fig:ConvGeoArch}. We utilise mean-pooled feature vectors without any attention-pooling or modules for refinement. The ConvNeXt is a modernised variant of the ResNet~\cite{he2016deep} and uses many improvement such as a patchified stem, higher kernel sizes, GELU activation functions, LayerNorm and depthwise convolutions. While current work does not use weight-sharing due to the different view-domains, we emphasise the use of a weight-shared ConvNeXt as a single encoder for both views. 
\subsection{Near Neighbour Sampling Based on GPS}
 In most of the cases hard examples can not be selected before training, since the model is not fitted to the domain of the training data. In order to bypass this drawback we leverage the geographical nature of cross-view geo-localisation. Images from the same area, or even the same street in the VIGOR dataset, share common properties like vegetation, street signs, or housing type. Since these are easy features that contrast a rural from a urban scenery, they will not contribute much to the objective function. Therefore we propose a simple GPS-based sampling strategy for the negative mining initialisation before the training even started. For CVUSA and VIGOR GPS locations are present in the dataset and near neighbours are selected based on the haversine distance. In the CVACT dataset the locations are in the UTM coordinate system, which is why we use the euclidean distance to determine the neighbours. The GPS coordinate is only used during training for the sampling strategy in the early epochs. For the University-1652 dataset this initialisation is not possible, therefore a simple shuffling is used. 
\subsection{Dynamic Similarity Sampling}
After a certain amount of epochs, GPS-based sampling is replaced by our Dynamic Similarity Sampling (DSS). During one full inference epoch on the training data the visual distances between all samples are calculated using the cosine similarity. In order to sample future batches the top $K$ nearest neighbours for every query image are selected. We set $K$ smaller or equal than the batch size, to have multiple regions or cities in one batch. These $K$ neighbours are sorted based on their similarity and then $\frac{k}{2}$ nearest samples are selected for the batch, the remaining $\frac{k}{2}$ samples are randomly selected from the remaining samples in $K$. The random selection process ensures enough diversity for the hard negatives since we only calculate new distances every $e$ epochs to shorten the training process. We set our hyperparameters as follows: $k=64$, $K=128$ and $e=4$. An ablation over the best choice of $k$ can be found in our supplementary material. Before we add our $k$ samples to the batch, a lookup is proceeded to avoid double entries within an epoch. For different settings of $k$ we did not observe the collapsing model problem described in~\cite{wu2017sampling} even when $k=K$. 

\section{Evaluation}
\subsection{Experiments and Results}
In our evaluation we conducted experiments on four standard benchmarks, namely CVUSA, CVACT, University-1652 and VIGOR. In the subsequent tables we compare our approach with previous work.
\subsubsection{CVUSA \& CVACT}
\paragraph{CVUSA} As one of the first cross-view datasets, CVUSA~\cite{workmann2015cvusa, zhai2017predicting} consists of 35,532 view pairs for training and 8,884 for evaluation and is one of the standard benchmarks. The satellite with a resolution of $750 \times 750$ and street-view images with a resolution of $224 \times 1232$ are aligned based on the camera extrinsics. This alignment ensures that the geographical north is located in the upper centre of the satellite image, and the street-view image is warped accordingly. The task here is a 1 to 1 mapping, every satellite-view has one corresponding street-view image. 
\paragraph{CVACT} In the dataset from Liu et al.~\cite{liu2019lending} the training and validation splits are the same size as CVUSA, but they provide an additional test split with $92802$ images. Unlike CVUSA, Canberra (Australia) is used as the region in the dataset and mostly urban scenery is included. Also a higher image resolution is provided with $1200 \times 1200$ for satellite and $832 \times 1664$ for street-views. The task is still a 1 to 1 mapping and the images are aligned like in the CVUSA dataset. For both CVACT and CVUSA, the metrics are Recall@k (R@k) with $k \in \{1,5,10\}$ and R@1\%. 

\paragraph{Results} As shown in table~\ref{tab:cvusaandcvact} our approach outperforms all previous work in the R@1 metric. Since the CVUSA and CVACT Val are saturated as a benchmark we also compare our work on the CVACT test set to show the generalisation ability of our model. The test split of the CVACT dataset is more difficult, because streetview images were sampled at a much higher density thus resulting in many semi-positive matches in the R@1 compared to R@5. It should also be noted that we do not use polar transformation. Approaches that do use the polar transformation usually experience a slight improvement on the benchmarks. As input image size for both datasets we use $140 \times 768$ for the street-view and $384\times 384$ for the satellite-view.
\begin{table*}[t!]
   \begin{center}
    \resizebox{\textwidth}{!}{ 
    \begin{tabular}{l|cccc|cccc|cccc} \hline \hline
       \multirow{2}{*}{Approach} & \multicolumn{4}{c|}{CVUSA} & \multicolumn{4}{c|}{CVACT Val}& \multicolumn{4}{c}{CVACT Test}  \\  
        & R@1 & R@5 & R@10  & R@1\% &  R@1 & R@5 & R@10  & R@1\%&  R@1 & R@5 & R@10  & R@1\%\\ \hline
       LPN~\cite{wang2021each} & 85.79 & 95.38 & 96.98 & 99.41 & 79.99 & 90.63 & 92.56 & - & - & - & - & - \\
       SAFA$^{\dagger}$~\cite{shi2019safa} & 89.84 & 96.93 & 98.14 & 99.64 & 81.03 & 92.80 & 94.84 & - & - & - & - & -\\
       TransGeo~\cite{zhu2022transgeo} & 94.08 & 98.36 & 99.04 & 99.77 & 84.95 & 94.14 & 95.78 & 98.37 & - & - & - & -\\
       GeoDTR~\cite{zhang2022cross} & 93.76 & 98.47 & 99.22 & 99.85 & 85.43 & 94.81 & 96.11 & 98.26 & 62.96 & 87.35 & 90.70 & 98.61\\ 
       GeoDTR$^{\dagger}$ & 95.43 & 98.86 & 99.34 & 99.86 & 86.21 & 95.44 & 96.72 &  98.77 & 64.52 & 88.59 & 91.96 & \textbf{98.74}\\ 
       SAIG-D~\cite{zhu2023simple} & 96.08 & 98.72 & 99.22 & 99.86 & 89.21 & 96.07 & 97.04 &  98.74 & - & - & - & -\\ 
       SAIG-D$^{\dagger}$~\cite{zhu2023simple} & 96.34 & 99.10 & 99.50 & 99.86 & 89.06 & 96.11 & 97.08 &  \textbf{98.89} & 67.49 & 89.39 & 92.30 & 96.80\\ 
       Ours & \textbf{98.68} & \textbf{99.68} & \textbf{99.78} & \textbf{99.87} & \textbf{90.81} & \textbf{96.74} & \textbf{97.48} & 98.77  & \textbf{71.51} & \textbf{92.42} & \textbf{94.45} & 98.70\\ \hline \hline
    \end{tabular}
    }
   \end{center}
     \caption{\textbf{Quantitative comparison between our approach and state-of-the-art approaches on CVUSA~\cite{zhai2017predicting} and CVACT~\cite{liu2019lending}.}  $\dagger$ denotes which models are using the polar transformation for their satellite input as a pre-processing technique.}
    \label{tab:cvusaandcvact}
\end{table*}

\subsubsection{University-1652}
While the previous two datasets matches $360^{\circ}$ street-views with satellite data, the dataset proposed by Zheng et al.~\cite{zheng2020university} includes multiple drone images from a building. The training images consists of over $50k$ drone views from $701$ university buildings and the task is to match the drone views to the according satellite image and vice versa. Another peculiarity of this dataset is that several buildings in the test set represent only adversary classes and there are no drone views referencing them, thus additional to the recall also the average precision (AP) is indicated. In our comparison~\ref{tab:uniS2D} we show the capability of your approach to different domains since we have a $1:n$ mapping. As input image size we downsample from $512 \times 512$ to $384 \times 384$ for both the UAV and the satellite-view.
\begin{table}[]
    \centering
   \begin{center}
    \resizebox{\columnwidth}{!}{ 
    \begin{tabular}{l|cc|cc} \hline \hline
        Approach &\multicolumn{2}{c|}{Drone2Sat}& \multicolumn{2}{c}{Sat2Drone} \\
         & R@1 & AP & R@1  & AP  \\ \hline \hline
        LPN~\cite{wang2021each} & 75.93 & 79.14 & 86.45 & 74.79\\
        SAIG-D~\cite{zhu2023simple} & 78.85 & 81.62 & 86.45 & 78.48\\
        DWDR~\cite{Wang2022LearningCG} & 86.41 & 88.41 & 91.30 & 86.02\\
        MBF~\cite{zhu2023uavworth} & 89.05 & 90.61 & 92.15 & 84.45 \\
        Ours & \textbf{92.65} & \textbf{93.81} & \textbf{95.14} & \textbf{91.39}\\ \hline \hline
    \end{tabular}
    }
   \end{center}
    \caption{\textbf{Quantitative comparison between our approach and state-of-the-art approaches on University-1652~\cite{zheng2020university}.}}
    \label{tab:uniS2D}
\end{table}
\subsubsection{VIGOR}
For VIGOR, Zhu et al.~\cite{zhu2021vigor} collect 90,618 satellite reference images and 105,214 street-view query images. The four cities, New York, Seattle, San Francisco and Chicago in this benchmark are used for two different split settings: same-area and cross-area. In the SAME setting, images of all cities are used in training and validation and in CROSS setting, training carried out on New York and Seattle and evaluation is done on San Francisco and Chicago to test the generalisation of the approach. Another novelty is the introduction of so-called semi-positive images. For each positive pair, there are three semi-positive sat-view neighbours which also cover regions of the street-view image. The position from which they were taken is not in the centre region of the image. Due to these three semi-positive images for each satellite image, it is difficult to achieve a high R@1 score. To determine the performance without semi-positive images, the hit rate is calculated. The hit rate is understood here as the R@1 with masked semi-positives, as these serve as a distraction. The images are provided with a resolution of $640 \times 640$ for satellite and $1024 \times 2048$ for street-views. We use $384 \times 768$ for the street-view and $384\times 384$ for the satellite-view as input size for our model.

In the same-area setting, table~\ref{tab:vigorComp}, we already outperform current work by far and show the ability of our approach even in the harder to handle CROSS setting. The higher hit rate and the performance at Recall@5 also shows that all approaches naturally have problems to distinguish between the semi-positives and the ground truth image as they are very close to each other. A visualisation of which features in an image make it difficult to distinguish between different ground truth and semi-positive images is provided in section~\ref{sec:visual}. While transfer to new, unknown cities and regions has been difficult in previous work, our model also shows outstanding performance in the cross-area setting. 

\begin{table}[]
    \centering
   \begin{center}
    \resizebox{\columnwidth}{!}{ 
    \begin{tabular}{l|cccccc} \hline \hline
       Approach & R@1 & R@5 & R@10  & R@1\% & Hit Rate \\ \hline
    \textbf{SAME} \\ 
        ~~SAFA$^{\dagger}$~\cite{shi2019safa} & 33.93 & 58.42 & 68.12 & 98.24 & 36.87\\
        ~~TransGeo~\cite{zhu2022transgeo} & 61.48 & 87.54 & 91.88 & 99.56 & 73.09\\
        ~~SAIG-D~\cite{zhu2023simple} & 65.23 & 88.08 & - & \textbf{99.68} & 74.11\\
        ~~Ours & \textbf{77.86} & \textbf{95.66} & \textbf{97.21} & 99.61 & \textbf{89.82}\\ \hline 
    \textbf{CROSS} \\ 
        ~~SAFA$^{\dagger}$~\cite{shi2019safa} & 8.20 & 19.59 & 26.36 & 77.61 & 8.85\\
        ~~TransGeo~\cite{zhu2022transgeo} & 18.99 & 38.24 & 46.91 & 88.94 & 21.21\\
        ~~SAIG-D~\cite{zhu2023simple} & 33.05 & 55.94 & - & 94.64 & 36.71\\
        ~~Ours & \textbf{61.70} & \textbf{83.50} & \textbf{88.00} & \textbf{98.17} & \textbf{69.87}\\ \hline \hline
    \end{tabular}
    }
    \end{center}
    \caption{\textbf{Quantitative comparison between our approach and the current state-of-the-art on VIGOR~\cite{zhu2021vigor}.} The same split contains images from all four cities in the dataset in training and validation. The cross split contains exclusively two cities and the model is then evaluated on the unknown other two cities. Our result show a much better generalisation on unknown regions than previous approaches. $\dagger$ denotes which models are using the polar transformation for their satellite input as a pre-processing technique.}
    \label{tab:vigorComp}
\end{table}

\subsection{Implementation Details}
In our experiments the architecture is not altered and the ConvNeXt-B with $88M$ parameters is used. An ablation of different architecture sizes and types is provided in section~\ref{sec:archchoice}. To minimise the overfitting on the training set we use label smoothing of $0.1$ in the InfoNCE loss and the temperature parameter $\tau$ is a learnable parameter. 
As Zhang et al.~\cite{zhang2022cross} showed, synchronous augmentations are important in order not to disturb the positions and orientations encoded in the image. Therefore, we used synchronous horizontal flipping as well as rotation on the satellite images and shift the street-view images accordingly to preserve the orientation of the satellite images. In order not to focus completely on certain regions in some images, we also use grid and coarse dropout and to further increase the generalisation we use colour jitter. Since University-1652 is a $1:n$ or $n:1$ task we use a custom sampler to prevent the occurrence of multiple images from the same class within the batch. The InfoNCE Loss treats all other examples as negatives, thus without a sampling strategy positive examples would be treated as negatives too. 
Each experiment is trained for 40 epochs with a batch size of 128 using the AdamW optimiser~\cite{loshchilov2017decoupled} with a initial learning rate of 0.001 and a cosine learning rate scheduler with a 1 epoch warm-up period.
\subsection{Ablation Study}
In our ablation study we dive into the design choices made in this work like the model architecture or the effectivity of our sampling. To provide additional insights we extract activation heatmaps of the satellite and street-view images of our model to visualise important regions in both views. 

\subsubsection{Architecture Evaluation}
\label{sec:archchoice}
For a long time, within the research community of cross-view geo-localisation, CNNs were the most important building block for learning useful representations~\cite{shi2019safa,hu2018cvm,zhai2017predicting,toker2021coming,zhang2022cross}. More recent publications explore other architectures such as transformers~\cite{zhu2022transgeo,yang2021l2ltr,zhao2022mutual} or the MLP mixer~\cite{zhu2023simple}. In order to determine a proper selection for our approach, we compare two standard architectures: a Vision Transformer (ViT)~\cite{dosovitskiy2020image} and a current state-of-the-art CNN~\cite{liu2022convnet} in table~\ref{tab:vigorArch}. 

The advantage of a CNN is that it can handle different input image sizes. Vision Transformers are not capable of dealing with multiple input sizes due to their fixed size positional encoding. Another constraint of the Transformer architecture is its scalability in terms of quadratic memory consumption due to the attention mechanism. Previous approaches propose to use separate encoders for satellite and street-view images. Therefore we also tested ConvNeXt without weight sharing, but we achieve better results when using the same encoder for both views. In table ~\ref{tab:vigorArch}, we compare models with two separate encoders for satellite and street-view with our approach that requires only one encoder for both views.

Since SAIG-D, one of the best approaches so far on datasets like VIGOR, CVUSA and CVACT, has a much smaller number of parameters a comparison is necessary. The question is how well our model performs when using the similar number of parameters. As shown in table~\ref{tab:vigorArch} the results from our work is very consistent, a higher amount of parameters has only a marginal impact on the performance. Instead of providing two individual backbones per view we use a single encoder and show it outperforms two separate encoders, even in settings with similar model size.

\begin{table}[]
    \centering
   \begin{center}
    \resizebox{\columnwidth}{!}{ 
    \begin{tabular}{l|ccccccc} \hline \hline
        Approach & \#Params (M) & Shared W. & R@1 & R@5 & R@1\% & Hit Rate \\ \hline \hline
        SAFA~\cite{shi2019safa} & $14.7 * 2$ & X & 33.93 & 58.42 & 98.24 & 36.87\\
        TransGeo~\cite{zhu2022transgeo} & $22.4 * 2$ & X & 61.48 & 87.54 & 99.56 & 73.09\\
        SAIG-D~\cite{zhu2023simple} & $15.6 * 2$ & X & 65.23 & 88.08 & 99.68 & 74.11\\
        Ours (ViT) & $86.8 * 2$ & X& 69.89 & 92.81 & 99.43 & 83.49\\
        Ours (Base) & $88.5* 2$ & X & 75.81 & 94.86 & 99.61 & 88.08\\\hline
        Ours (Nano) & 15.5 & \checkmark & 70.70 & 91.17 & 94.19 & 99.40\\
        Ours (Tiny) & 28.5 & \checkmark & 73.90 & 93.60 & 99.47 & 84.77\\
        Ours (Base) & 88.5 & \checkmark & 77.86 & 95.66 & 99.61 & 89.82\\
        Ours (Large) & 197.7 & \checkmark & 78.27 & 95.98 & 99.66 & 90.50 \\\hline \hline
    \end{tabular}
    }
   \end{center}
    \caption{\textbf{Parameter efficiency of our models compared to the current state-of-the-art.} Increasing the number of parameters brings only a marginal advantage and shows that our sampling and the symmetric loss is more important. Results are reported on the VIGOR SAME split.}
    \label{tab:vigorArch}
\end{table}

\subsubsection{Effectiveness of Sampling Strategies}

Without any sampling strategy we still achieve competitive results when using contrastive training leveraging the InfoNCE loss. But a good sampling strategy drastically improves the performance, as shown in table~\ref{tab:vigorSampling}. We also tested the sampling strategies for CVUSA and CVACT, and the results are included in our supplementary material.

Due to the presence of geographical neighbours in the VIGOR SAME split, the model benefits from this method during the whole training course. However, since locations from all four cities can occur in the SAME split, an ablation is also interesting to see whether GPS sampling has an effect on the cross area split. Those nearby locations are not present in the test set and an increase in performance with the help of GPS sampling on the cross-area split would indicate the generalisation of this sampling. As we successfully show the results are consistent in our same-area and cross-area experiments. Even if only GPS sampling is used during the training we achieve almost the same R@1 as with visual similarity sampling. When using GPS-based sampling without DSS, it can be assumed, that areas within short geographical distances are also quite similar and harder to distinguish based on visual features. The performance with a combination of both strategies is larger in the cross-area setting in comparison to the same-area setting. Based on this analysis we decided to use both sampling strategies for the reported CVUSA and CVACT results. 

\begin{table}[]
    \centering
   \begin{center}
    \resizebox{\columnwidth}{!}{ 
    \begin{tabular}{l|cccccc} \hline \hline
       Sampling & R@1 & R@5 & R@10  & R@1\% & Hit Rate \\ \hline
    \textbf{SAME} \\ 
        ~~Random & 65.23 & 91.62 & 95.85 & 99.85 & 78.77\\
        ~~GPS & 74.69 & 92.21 & 94.66 & 99.34 & 83.49\\
        ~~DSS & 76.94 & 95.50 & 97.12 & \textbf{99.66} & 89.31\\
        ~~GPS~+~DSS &\textbf{77.86} & \textbf{95.66} & \textbf{97.21} & 99.61 & \textbf{89.82}\\ \hline 
    \textbf{CROSS} \\ 
        ~~Random & 36.38 & 63.96 & 72.43 & 97.18 & 43.66\\
        ~~GPS & 57.17 & 78.44 & 83.79 & 97.22 & 62.81\\
        ~~DSS & 58.59 & 81.45 & 86.44 & 97.98 & 66.63\\
        ~~GPS~+~DSS &\textbf{61.70} & \textbf{83.50} & \textbf{88.00} & \textbf{98.17} & \textbf{69.87}\\ \hline \hline
    \end{tabular}
    }
    \end{center}
    \caption{\textbf{Quantitative comparison between our proposed sampling techniques for the VIGOR SAME split.}}
    \label{tab:vigorSampling}
\end{table}

\subsubsection{Generalisation Capabilities}
\begin{table}[]
    \centering
   \begin{center}
    \resizebox{\columnwidth}{!}{ 
    \begin{tabular}{l|cccccc} \hline \hline
       Approach & R@1 & R@5 & R@10  & R@1\%  \\ \hline
    \textbf{CVUSA $\xrightarrow{}$ CVACT} \\ 
        ~~L2LTR~\cite{yang2021l2ltr} & 47.55 & 70.58 & - & 91.39 \\
        ~~L2LTR$^{\dagger}$~\cite{yang2021l2ltr} & 52.58 & 75.81 & - & 93.51 \\
        ~~GeoDTR~\cite{zhang2022cross} & 47.79 & 70.52 & - & 92.20 \\
        ~~GeoDTR$^{\dagger}$ & 53.16 & 75.62 & - & 93.80 \\
        ~~Ours &\textbf{56.62} & \textbf{77.79} & \textbf{87.02} & \textbf{94.69} \\ \hline 
    \textbf{CVACT $\xrightarrow{}$ CVUSA} \\
        ~~L2LTR~\cite{yang2021l2ltr} & 33.00 & 51.87 & - & 84.79 \\
        ~~L2LTR$^{\dagger}$~\cite{yang2021l2ltr} & 37.69 & 57.78 & - & 89.63 \\
        ~~GeoDTR~\cite{zhang2022cross} & 29.13 & 47.86 & - & 81.09 \\
        ~~GeoDTR$^{\dagger}$ & 44.07 & \textbf{64.66} & - & 90.09 \\
        ~~Ours &\textbf{44.95} & 64.36 & \textbf{72.10} & \textbf{90.65} \\ \hline \hline
    \end{tabular}
    }
    \end{center}
    \caption{\textbf{Generalisation ablation when trained on the CVUSA dataset and evaluated on CVACT and vice versa.} $\dagger$ denotes approaches that used the polar transformation.}
    \label{tab:generalisation}
\end{table}
A very interesting question about different approaches is the generalisation capability, i.e. the ability of models trained on one region or scenery and then used for another region. As we have already shown in table~\ref{tab:vigorComp} in particular, our approach generalises to unknown cities very well, but not yet with nearly the same performance as in the same-area setting. For our ablation, we additionally evaluated the models trained on CVUSA and CVACT to show the transfer between these datasets. As can be seen from table~\ref{tab:generalisation}, our approach scores well, especially when compared to approaches without polar transformation. In our supplementary material we provide further visualisations of the behaviour of our model for this setting. This shows that incorrectly recognised satellite images are usually very similar in terms of the course of the road. The model is biased towards the road course in the image and encodes it in the feature vector. This is one of the reasons why previous approaches benefit from polar transformation, which leads to a better alignment of street-view and distorted satellite scenery.

\subsubsection{Visualisation}
\label{sec:visual}
\begin{figure}[t]
    \centering
    \includegraphics[width=0.\columnwidth]{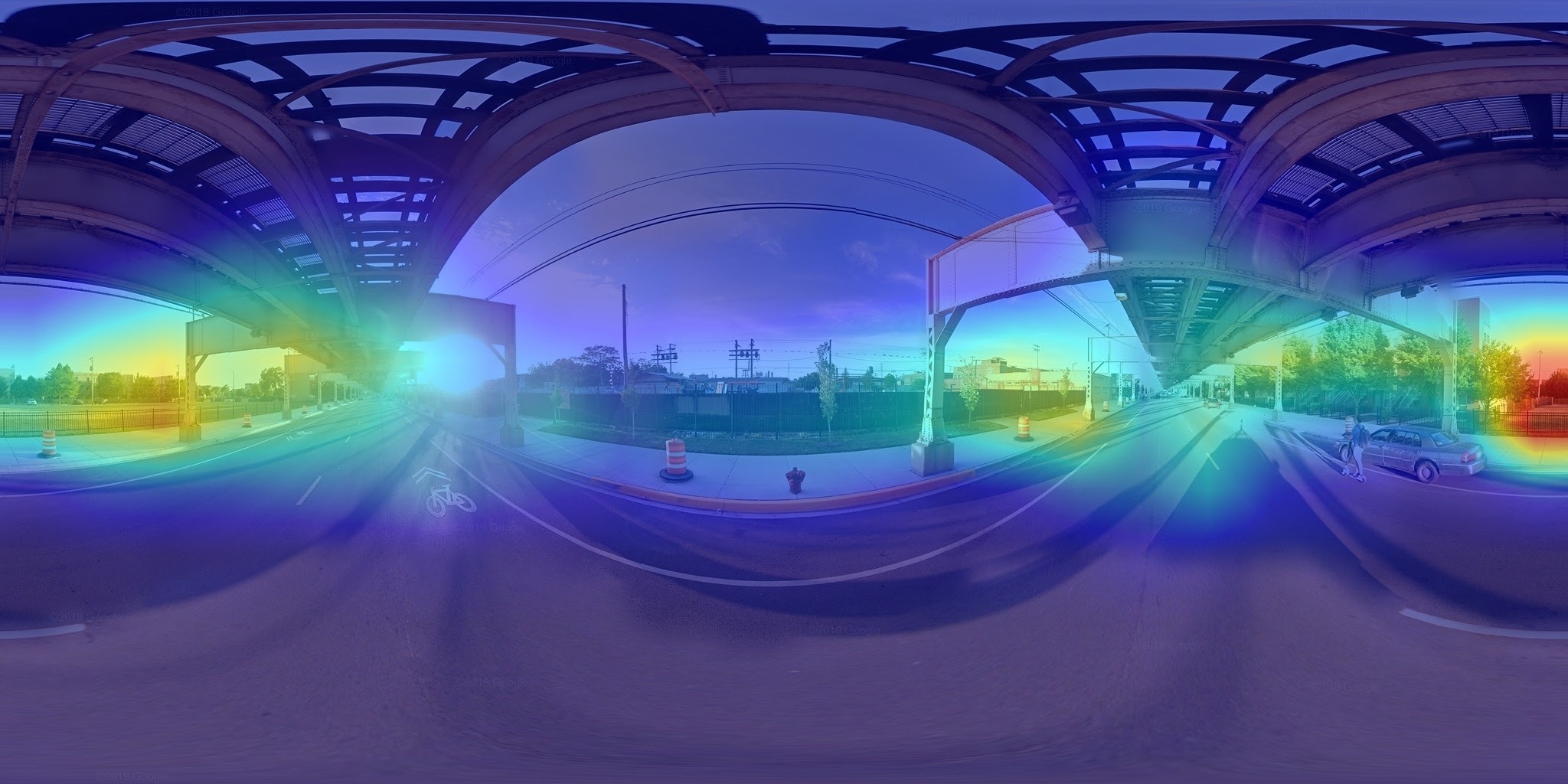}

    \includegraphics[width=0.495\columnwidth]{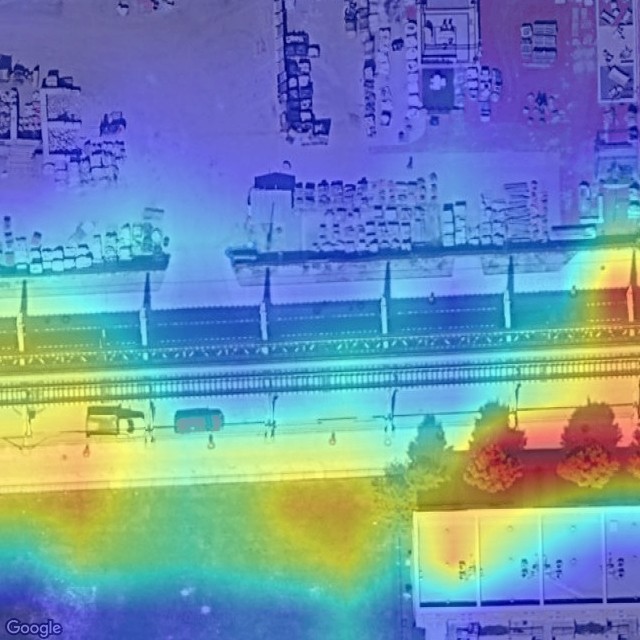}
    \hfill
    \includegraphics[width=0.495\columnwidth]{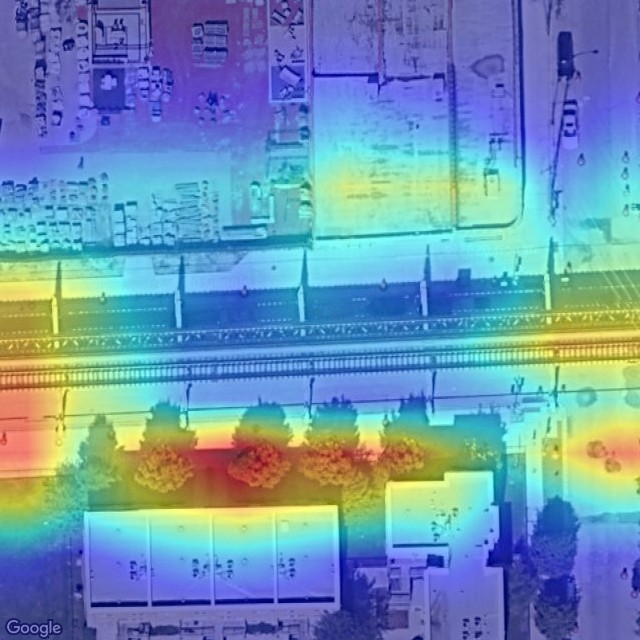}
    \caption{\textbf{Heatmap for a semi-positive hit on the VIGOR dataset.} The upper picture shows the street-view query, the bottom left side the prediction and on the right side the actual ground truth is shown. Features like trees and buildings are fundamental for the prediction and in both similar features are relevant.}
    \label{fig:vigor}
\end{figure}
Human reasoning for cross-view geo-localisation is usually based on powerful features such as the course of a road or distinctive buildings. Therefore, it is particularly interesting to understand which regions are important for the model. In our supplementary material we present multiple examples from CVUSA. These examples show that in CVUSA streets are one of the most important features in an image. For the VIGOR dataset this is not the case. In figure~\ref{fig:vigor} a semi-positive hit can be observed and activations from the surrounding landscape are more important for the prediction. However, the VIGOR dataset contains multiple semi-positives for each position. A clear prediction is harder since multiple important features, like the trees in figure~\ref{fig:vigor} are present in both satellite images.
\section{Conclusion}

In our work, we present a simple but effective approach to solve the geo-localisation task. Our proposed model consists of a single image encoder for both satellite and ground view based on a modern CNN. This lightweight approach leverages contrastive learning by using the InfoNCE Loss as training objective. We further demonstrate, that an effective sampling strategy leads to superior results on the VIGOR, CVUSA, CVACT and University-1652 datasets. Compared to other state-of-the-art approaches, our model does not need complex pre-processing steps or any task specific aggregation modules, nor the use of loss surface smoothing like SAM or ASAM to achieve outstanding generalization performance in cross-area settings. 

\section{Discussion}
During our work we identified several problems as well as future challenges. While the traditional datasets such as CVUSA, CVACT and VIGOR only contain $360^{\circ}$ street-view images, University-1652 is one of the few datasets that contain a limited field of view of the drone images. However, all images are taken close to the building. This simplifies the geo-localisation task enormously in both cases as non-orientation bound features have to be matched. Another reason for the outstanding performance on CVUSA and CVACT is the perfect alignment of the position of the Street View image to the centre of the satellite image. This is fixed in VIGOR, but as well here the benchmark becomes progressively saturated. To improve the applicability, additional datasets with explicitly not aligned central positions and unknown geographical orientation between satellite and street-view image are required. In addition, the FoV should be limited to below 120 degrees, as this reflects the FoV from a current standard smartphone.

Another common characteristic of the previous datasets is the focus on urban environments. CVUSA offers a somewhat broader spectrum here, but the ground view images are obviously all taken on streets. This does not reflect the variety of scenes in the wild and as shown in our supplementary our approach focuses most times on the streets and intersections to achieve matching, especially on the CVUSA dataset. Future datasets should also include ground views not taken exclusively from roads to increase diversity, usefulness and practicability.

{\small
\bibliographystyle{ieeetr}
\bibliography{egbib}
}
\setcounter{section}{0}
\setcounter{figure}{0}

\section{Supplementary Material}

\section{Hyperparameter}
In order to determine the best hyperparameter for our Dynamic Similarity Sampling (DSS) we choose different $k$ on the VIGOR SAME split during sampling, where $k$ is the actually selected amount of neighbours from the nearest neighbour pool of size $K$. As can be seen in table~\ref{tab:vigorNeighbourComp} our approach performs best with $k=64$ and also does not collapse to simple solutions, when all hard negatives are sampled if $k=K$ without a random factor. 
\begin{table}[h!]
    \centering
   \begin{center}
    \resizebox{\columnwidth}{!}{ 
    \begin{tabular}{l|ccccccc} \hline \hline
        Neighbours $k$ & R@1 & R@5 & R@10  & R@1\% & Hit Rate \\ \hline \hline
        16 & 76.84 & 95.24 & 96.98 & 99.68 & 88.55\\
        32 & 77.39 & 95.61 & 97.19 & 99.65 & 89.46\\
        64 & \textbf{77.86} & \textbf{95.66} & \textbf{97.21} & \textbf{99.61} & \textbf{89.82}\\
        128 & 77.84 & 95.42 & 97.07 & 99.58 & 89.69 \\\hline \hline
    \end{tabular}
    }
   \end{center}
    \caption{Comparison between the different neighbours we select during our similarity sampling on the VIGOR same split, with pool size $K=128$. Based on this analysis we decided to use $k=64$ for all our experiments.}
    \label{tab:vigorNeighbourComp}
\end{table}
\section{Symmetric InfoNCE Loss}
In the analysis of our approach, we also examined to what extent the symmetrical InfoNCE loss contributes to the performance. As shown in table~\ref{tab:symInfoNCE}, the direction of the loss calculation plays a significant role. The performance suffers, when the similarity of the satellite-view as query is calculated against the street-view as reference. Whereas if we use the street-view as query and the satellite-view as reference, the performance is almost on par with the symmetric loss calculation in both directions. An explanation for this behaviour comes from the data in VIGOR and our similarity sampling. Since in our DSS we are looking for street-view images whose distance is minimal to satellite images, the other direction cannot benefit from our sampling. To avoid this and to benefit from both directions, we formulated the loss symmetrically to achieve the best performance.
\begin{table}[]
    \centering
   \begin{center}
    \resizebox{\columnwidth}{!}{ 
    \begin{tabular}{l|ccccccc} \hline \hline
        Loss Direction & R@1 & R@5 & R@10  & R@1\% & Hit Rate \\ \hline \hline
        Sat $\rightarrow{}$ Street& 74.69 & 93.80 & 95.81 & 99.40 & 86.27\\
        Street $\rightarrow{}$ Sat & 77.49 & 95.66 & 97.18 & 99.58 & 89.57\\
        Street $\leftrightarrow$ Sat & \textbf{77.86} & \textbf{95.66} & \textbf{97.21} & \textbf{99.61} & \textbf{89.82} \\\hline \hline

    \end{tabular}
    }
   \end{center}
    \caption{Comparison between unidirectional loss calculation and symmetric loss caluclation on the VIGOR same split. Our model profits from a symmetric loss the most.}
    \label{tab:symInfoNCE}
\end{table}
\section{Loss comparison}
The triplet loss is very prone to model collapsing when using only hard negatives within a batch. A proposed extension to overcome this is the soft-margin triplet loss. In Table~\ref{tab:triplet} we compare the two triplet losses with the InfoNCE loss. As we show the model collapses when using the triplet loss without extension.
\begin{table}[h]
    \centering
   \begin{center}
    \resizebox{0.9\columnwidth}{!}{ 
    \begin{tabular}{l|ccccc} \hline \hline
       Dataset & R@1 & R@5 & R@10  & R@1\%  \\ \hline
        Triplet Loss & 0.00 & 0.06 & 0.10 & 1.26 \\  
        Soft-Margin Triplet & 91.83 & 97.87 & 98.75 & 99.67 \\ 
        InfoNCE &\textbf{98.68} & \textbf{99.68} & \textbf{99.78} & \textbf{99.87}\\\hline
    \end{tabular}
    }
    \end{center}
    \caption{\textbf{Loss function comparison for CVUSA.}}
    \label{tab:triplet}
\end{table} 

\section{Sampling Strategies}
We additionally perform the comparison of the sampling strategies on the CVUSA and CVACT dataset and come to similar results as for VIGOR. Our experiments show that regardless of the dataset, the combination of GPS sampling and DSS leads to the best results. Considering the two sampling strategies in isolation leads to different results, which strongly depend on the dataset. When the samples are from geographically close areas, as in CVACT (area around Canberraast) and VIGOR (four different cities), GPS sampling performs similarly to DSS. When there are a large number of distant locations, as in CVUSA, GPS sampling is no more advantageous than random sampling.
\begin{table}[]
    \centering
   \begin{center}
    \resizebox{0.9\columnwidth}{!}{ 
    \begin{tabular}{l|ccccc} \hline \hline
       Sampling & R@1 & R@5 & R@10  & R@1\%  \\ \hline
    \textbf{CVUSA} \\ 
        ~~Random & 97.83 & 99.63 & 99.75 & 99.89\\
        ~~GPS & 97.83 & 99.53 & 99.72 & 99.87 \\
        ~~DSS & 98.51 & 99.69 & 99.79 & 99.85\\
        ~~GPS~+~DSS &\textbf{98.68} & \textbf{99.68} & \textbf{99.78} & \textbf{99.87} \\ \hline 
    \textbf{$\text{CVACT}_{\text{test}}$} \\ 
        ~~Random & 60.57 & 89.50 & 92.99 & 98.92 \\
        ~~GPS & 71.13 & 90.28 & 92.47 & 98.09 \\
        ~~DSS & 71.04 & 92.26 & 94.33 & 98.58 \\
        ~~GPS~+~DSS &\textbf{71.51} & \textbf{92.42} & \textbf{94.45} & \textbf{98.70}\\\hline
    \end{tabular}
    }
    \end{center}
    \caption{\textbf{Comparison of sampling for CVUSA/CVACT.}}
    \label{tab:cvusa}
\end{table}

\section{Visualisation}
To provide a better visualisation for the alignment of the activation maps we additionally apply a inverse polar transformation to the activation maps of the street view image. With this transformation it is easier to visualise the correspondence between both views more clearly. 

\subsection{Generalisation CVACT $\rightarrow$ CVUSA}
\begin{figure*}
    \centering
    \includegraphics[width=0.4\textwidth]{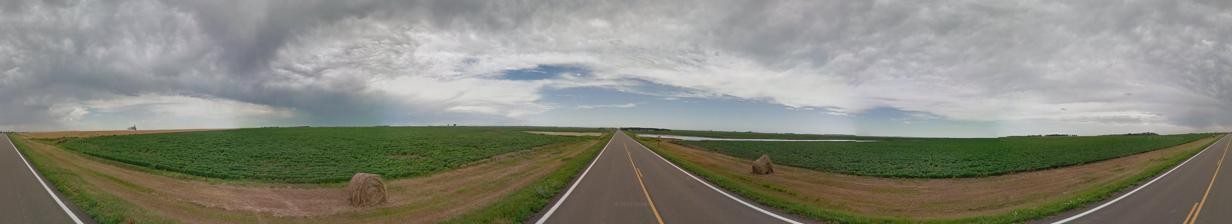} \includegraphics[width=0.4\textwidth]{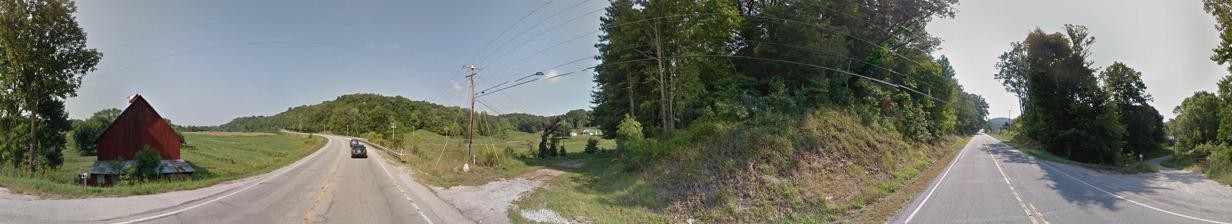} \vfill
    \includegraphics[width=0.4\textwidth]{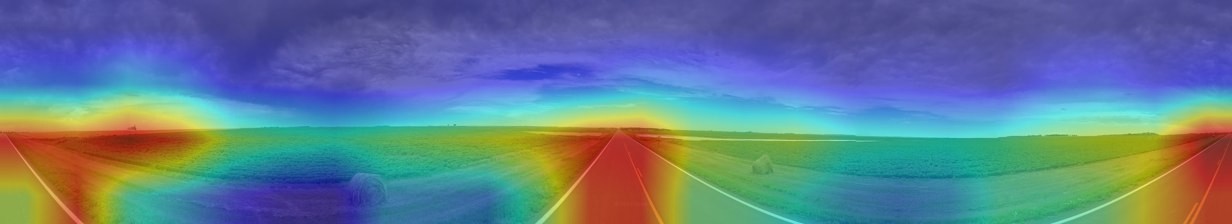} \includegraphics[width=0.4\textwidth]{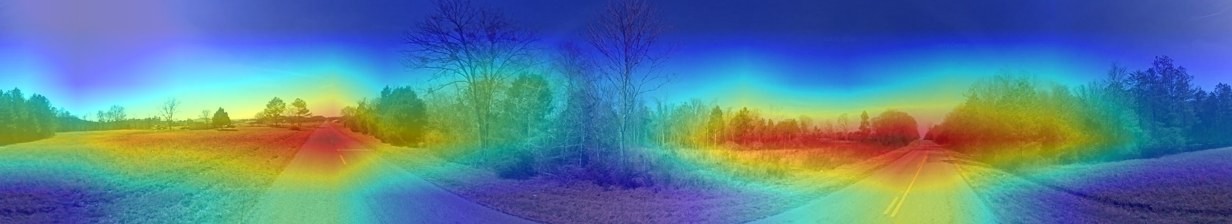}\vfill
    \subfloat[GT: similarity 0.4311]{
    \includegraphics[width=0.4\columnwidth]{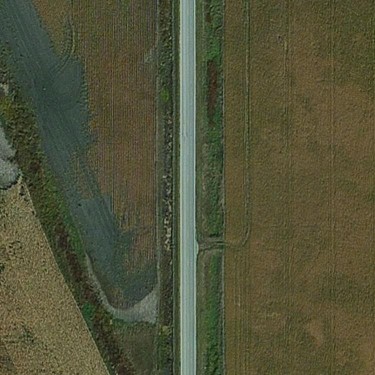}
    }
    \subfloat[Pred: similarity 0.670]{
    \includegraphics[width=0.4\columnwidth]{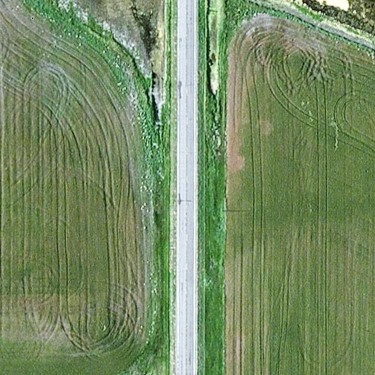}
    }
    ~
    \subfloat[GT: similarity 0.4708]{
    \includegraphics[width=0.4\columnwidth]{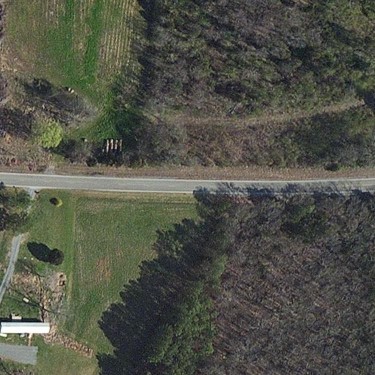}
    }
    \subfloat[Pred: similarity 0.4965]{
    \includegraphics[width=0.4\columnwidth]{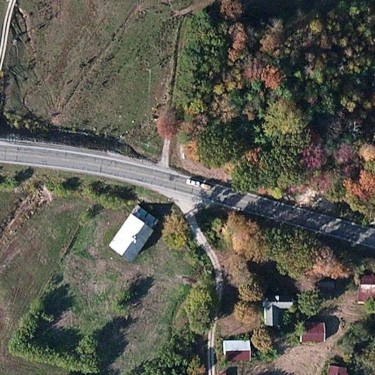}
    }
    \caption{\textbf{False predictions from the CVUSA dataset predicted with a model trained on CVACT.} In the left image the street course is very similar, but the street-view image seems to be taken at another season than the satellite image. On the left the course of the road differs but the vegetation matches better.}
    \label{fig:cvact2cvusa}
\end{figure*}

As we have shown in our previous results, our model also performs well on new regions compared to previous work. However, this performance is lower as when the same region is used, were the model is trained on. For this transferability we provide some insights for a model trained on CVACT predicting on CVUSA. We used randomly selected false predictions and compare them with each other visually as well as by the cosine similarity. The results are shown in~\ref{fig:cvact2cvusa} for two examples. 

\subsection{Correct predictions}
For VIGOR, our model has learned to pay attention particularly to vegetation and also on road markings, as shown in Figure~\ref{fig:vigor} for a correct predicted sample. 

\begin{figure*}
    \centering

    \includegraphics[width=\textwidth]{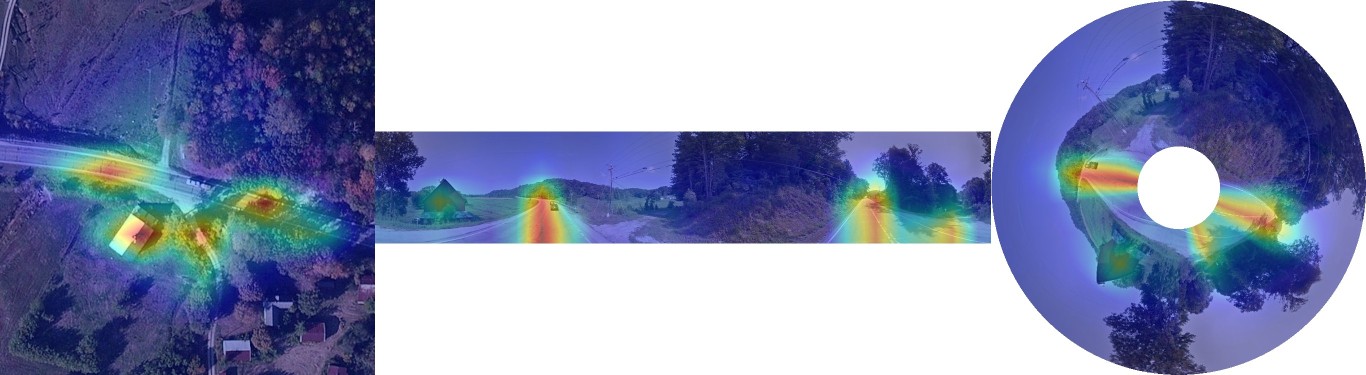}
    \includegraphics[width=\textwidth]{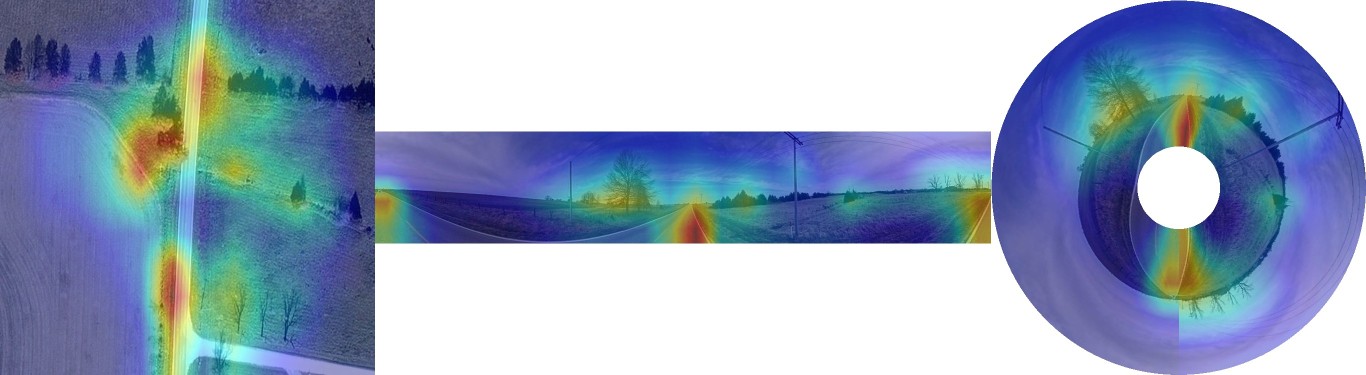}
    \includegraphics[width=\textwidth]{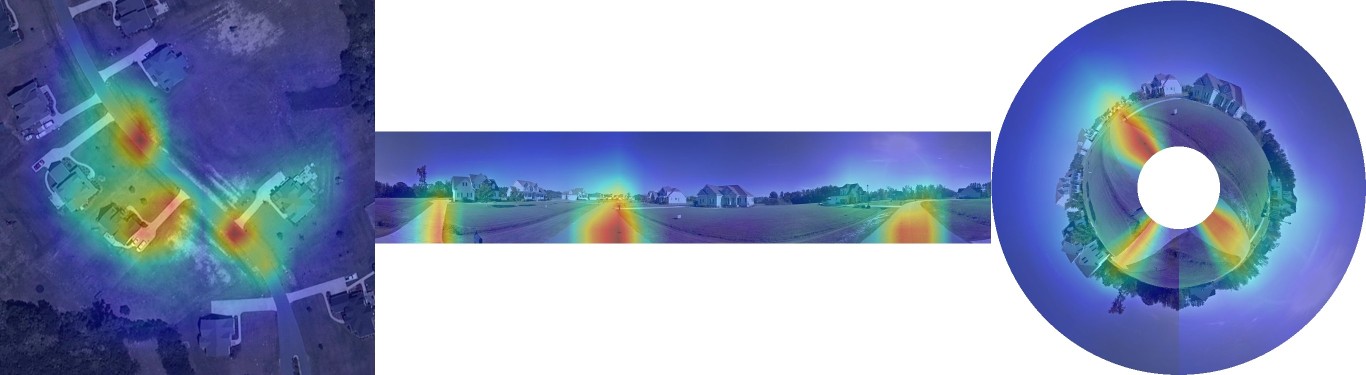}
    \includegraphics[width=\textwidth]{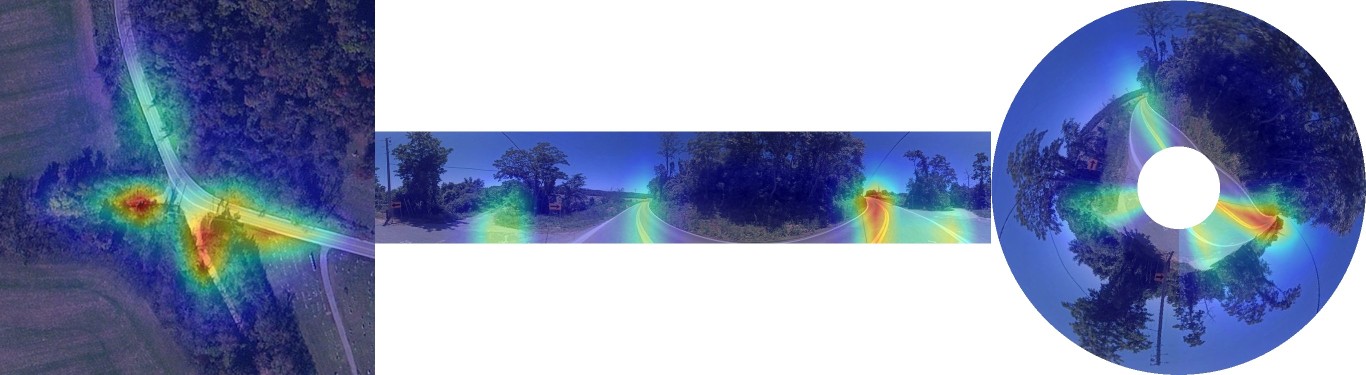}
    \caption{\textbf{Heatmap visualisations for correct predictions on the CVUSA dataset.} We use the inverse polar transformation to show the correspondence of the activation maps in satellite and street-views. According to the activation maps the most significant regions are the street course and position of intersections on that samples.}
    \label{fig:cvusa_withwarp}
\end{figure*}

We visualise the activation maps for some correct predictions from a in-domain trained model on CVUSA in figure~\ref{fig:cvusa_withwarp}. Here, the models mainly focus on the road features such as the course and the position of intersections. 

VIGOR covers more urban locations in its training data compared to CVUSA, so in rural settings the road itself plays a more dominant role. 

\section{Acknowledgement}
The authors gratefully acknowledge the computing time granted by the Institute for Distributed Intelligent Systems and provided on the GPU cluster Monacum One at the University of the Bundeswehr Munich.

\begin{figure*}
    \centering

    \includegraphics[width=0.4\textwidth]{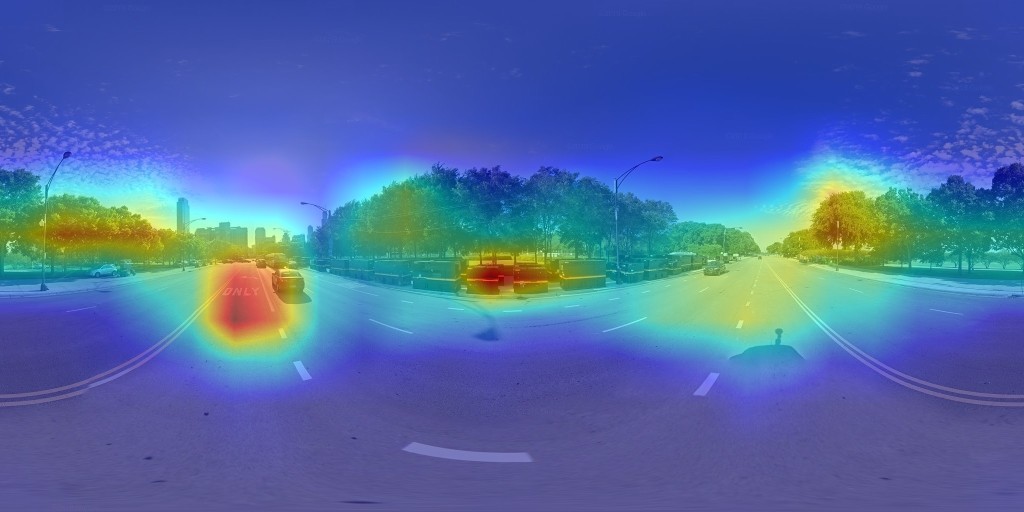} 
    \includegraphics[width=0.4\textwidth]{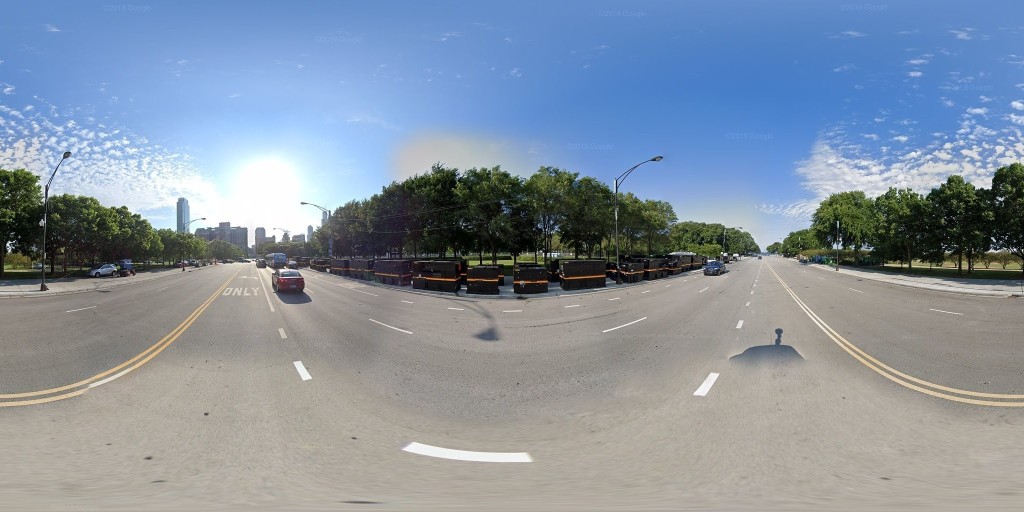} \vfill
    
    \includegraphics[width=0.4\textwidth]{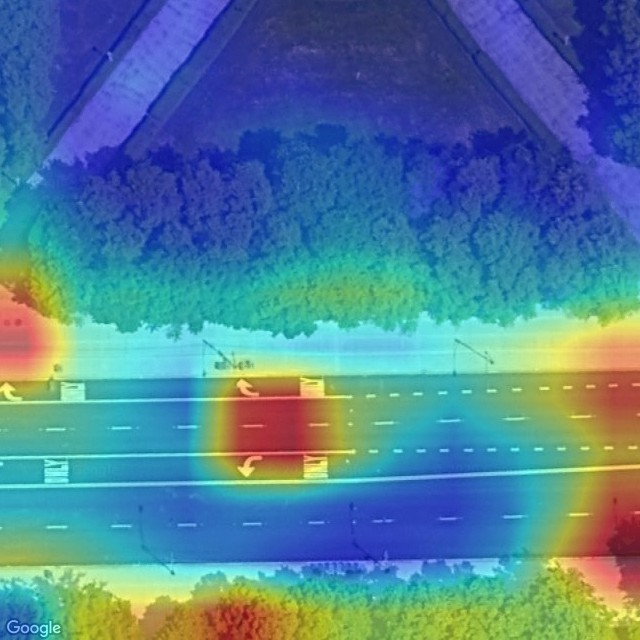}
    \includegraphics[width=0.4\textwidth]{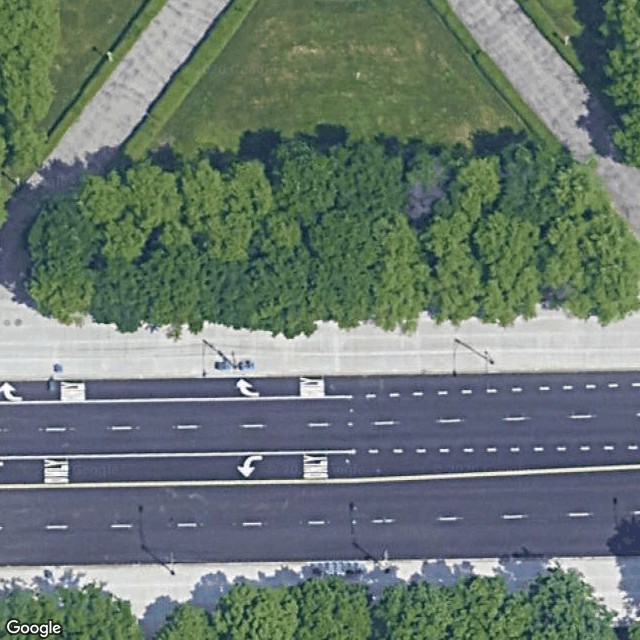}
    \caption{\textbf{Heatmap visualisation of a correct prediction on the VIGOR SAME split.} The model focuses here mainly on the road markings and the vegetation on both sides of the street.}
    \label{fig:vigor}
\end{figure*}

\section{Acknowledgement}
The authors gratefully acknowledge the computing time granted by the Institute for Distributed Intelligent Systems and provided on the GPU cluster Monacum One at the University of the Bundeswehr Munich.

\end{document}